\documentclass[runningheads,a4paper]{llncs}

\usepackage[T1]{fontenc}
\usepackage{graphicx}
\graphicspath{{figures/}}
\usepackage{amsmath,amssymb,amsfonts}
\usepackage{booktabs}
\usepackage{multirow}
\usepackage{makecell}
\usepackage{url}
\usepackage{xcolor}
\usepackage[hidelinks]{hyperref}

\begin{document}

\title{FENAS: A Few-Shot Meta-Ensembling Framework for Efficient Neural Architecture Search with Synergized Zero-Cost Proxies}

\titlerunning{FENAS}

\author{
Yifeng Mao\inst{1} \and
Yuze Feng\inst{1} \and
Yutong Shen\inst{1} \and
Liyang Gao\inst{1} \and
Zhixiang He\inst{1}\textsuperscript{*}
}

\authorrunning{Y. Mao et al.}

\institute{
Beijing University of Technology, Beijing, China\\
\email{maoyifeng@emails.bjut.edu.cn, zxhe@bjut.edu.cn} \textsuperscript{*}Corresponding author
}

\maketitle

\begin{abstract}
Efficient neural architecture search (NAS) requires accurate performance estimation under limited evaluation budgets. Existing zero-cost proxies (ZCPs) provide fast training-free signals, but each proxy usually captures only a partial property of an architecture, leading to unstable rankings across datasets and search spaces. In contrast, GNN-based predictors can improve accuracy, but often depend on labeled architectures, graph encoders, and costly representation learning, which makes them less suitable for extreme few-shot settings. We propose FENAS, a topology-agnostic few-shot prediction framework that turns complementary initialization priors into robust ranking signals. Its core component, CAST, is a feature pipeline that combines copula alignment and synergy transformation. CAST maps heterogeneous ZCPs into a compact feature manifold, enabling lightweight meta-learners to produce reliable rankings from scarce supervision. On NAS-Bench-201, with only 0.5\% labeled samples, FENAS achieves a Kendall's Tau ($\tau$) of 0.70, outperforming representative topology-based few-shot predictors in this low-label regime. In closed-domain search, FENAS evaluates the NAS-Bench-201 space in approximately 1138 seconds and achieves competitive test accuracy. By providing a fast, reliable, and structurally simple evaluation mechanism, FENAS supports the design of automated neural architectures under limited evaluation budgets.

\end{abstract}

\section{Introduction}

\begin{figure}[t]
  \centering
  \includegraphics[width=0.97\textwidth]{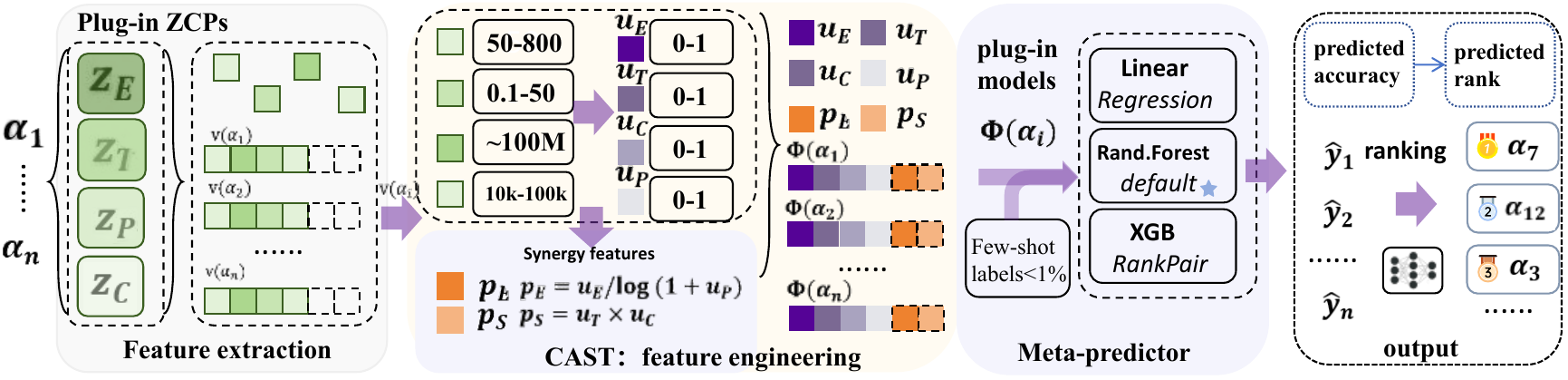}
  \caption{The overall framework of FENAS. The pipeline consists of three decoupled stages: (1) \textbf{Feature Extraction:} raw zero-cost proxy scores are extracted from complementary physical dimensions. (2) \textbf{CAST Feature Engineering:} heterogeneous proxy scores are aligned and transformed into a compact feature manifold. (3) \textbf{Meta-predictor:} a lightweight meta-learner uses extremely few-shot labels to output accurate performance rankings.}
  \label{fig:framework}
\end{figure}

Neural Architecture Search (NAS)~\cite{elsken2019neural} has become an important technique for automatically designing deep neural network architectures. It has achieved strong results in computer vision and natural language processing and provides a scalable way to adapt architectures to different tasks, datasets, and deployment constraints. However, conventional NAS methods usually require training or partially training a large number of candidate architectures, which leads to substantial computational overhead. Therefore, accurate performance estimation under extremely limited evaluation budgets has become a central challenge for efficient NAS~\cite{white2021bananas}.

Zero-Cost Proxies (ZCPs) offer a promising solution by estimating architecture quality at initialization from a single forward or backward pass. They can quickly screen weak candidates before expensive training and thus greatly reduce search cost. However, individual proxies are often unstable because each one captures only a limited physical property of an architecture, such as expressivity, trainability, or connectivity. As a result, their rank correlations vary significantly across datasets, tasks, and search spaces~\cite{ning2021evaluating,white2021powerful}, making raw ZCPs difficult to use as universal and reliable performance indicators.

Few-shot predictors based on Graph Neural Networks (GNNs) or other topology encoders~\cite{ning2020generic,xu2021renas,chen2021contrastive} provide another way to reduce evaluation cost. By explicitly modeling architecture graphs and learning topology-aware representations, these methods can improve prediction quality under limited supervision. Nevertheless, they introduce additional overhead from graph construction, message passing, and structured inference. More importantly, in extreme few-shot regimes, e.g., less than $1\%$ labeled samples, learning high-dimensional topology representations from scratch can be sample-inefficient and prone to overfitting. Our experiments show that on topology-stable benchmarks such as NAS-Bench-201, a lightweight topology-agnostic predictor can already outperform heavy topology-based methods; with only 78 labeled architectures, FENAS achieves a $\tau$ of 0.695 on CIFAR-10, surpassing DCL-ENAS~\cite{zhang2025evolutionary} (0.610) and CAP~\cite{ji2024cap} (0.600).

To address the tension between the instability of individual ZCPs and the complexity of topology-based predictors, we propose FENAS (Few-shot meta-Ensembling Neural Architecture Search), an efficient and topology-agnostic performance prediction framework. Our key insight is that, instead of spending model capacity on explicit topology learning, it is more effective to directly model the intrinsic priors of architectures and learn how to combine them with minimal supervision. Specifically, we extract features from four complementary physical dimensions: expressivity, trainability, connectivity, and parameter capacity. To transform these heterogeneous signals into a compact and learnable feature manifold, we design the CAST (Copula Alignment and Synergy Transformation) operator. Based on this representation, a lightweight meta-learner only needs a very small number of labeled architectures to capture the task-dependent synergy among these priors. Extensive experiments show that FENAS achieves strong ranking quality with very limited labeled architectures while avoiding explicit topology representation learning. In end-to-end search, FENAS remains competitive under constrained budgets,  completing the closed-domain search on NAS-Bench-201 in only $\sim$1138 seconds and remaining in the most efficient tier on DARTS with 0.3 GPU days.

The main contributions of this paper are as follows:
\begin{itemize}
    \item \textbf{We propose a lightweight topology-agnostic NAS prediction framework with strong efficiency advantages.} FENAS bypasses expensive graph construction and message passing, and instead performs ranking with multi-dimensional priors extracted at initialization. Under extremely limited supervision, it demonstrates strong sample efficiency and competitive search performance, while keeping the overall pipeline structurally simple and computationally lightweight.
    
    \item \textbf{We design CAST to synergize heterogeneous physical priors.} FENAS systematically integrates four complementary dimensions of architecture dynamics: expressivity, trainability, connectivity, and capacity. Through distribution alignment and synergy transformation, CAST mitigates scale inconsistency and improves the usability of raw ZCP signals for few-shot prediction.
    
    \item \textbf{We show that minimal supervision can turn unstable zero-cost signals into robust predictors.} Extensive experiments demonstrate that FENAS consistently improves over individual ZCPs and achieves particularly strong results in extreme few-shot settings. The interpretability analysis further shows that FENAS adaptively adjusts the importance of different physical dimensions across tasks, providing a direct explanation for its improved robustness and cross-dataset generalization.
\end{itemize}
\section{Related Work}

\subsection{Neural Architecture Search (NAS)}
Neural Architecture Search (NAS) aims to automate the design of high-performing neural networks, but its practical use is often limited by the high cost of evaluating candidate architectures in large search spaces. Early NAS methods based on reinforcement learning and evolutionary algorithms~\cite{zoph2017neural,real2019regularized} typically require training or partially training many architectures, leading to substantial computational overhead. To improve efficiency, differentiable and low-cost search strategies~\cite{liu2018darts} have been proposed to reduce the reliance on exhaustive evaluation. In parallel, benchmark suites such as NAS-Bench-201~\cite{Dong2020NAS-Bench-201:} provide standardized search spaces with ground-truth performance records, enabling controlled analysis of efficient NAS algorithms.

\subsection{Zero-Cost Proxies (ZCPs)}
Zero-Cost Proxies (ZCPs) estimate architecture quality at initialization using statistics from a single forward or backward pass~\cite{abdelfattah2021zerocost}. Existing proxies characterize architectures from different physical perspectives. Gradient and pruning-based methods, such as SynFlow~\cite{tanaka2020pruning}, measure parameter sensitivity or information flow, while activation-based methods such as NASWOT~\cite{mellor2021neural} capture expressivity through feature-map statistics or activation patterns. Although these proxies provide useful training-free priors, each individual proxy usually reflects only one local property of an architecture, making its ranking reliability highly dependent on the dataset and search space.

\subsection{Few-Shot Prediction and Proxy Fusion}
Few-shot NAS predictors aim to reduce the number of labeled architectures required for accurate performance estimation. Recent methods usually rely on graph topology encoding, self-supervised learning, or supernet-based mechanisms~\cite{yan2020does,wen2020neural,lu2021tnasp,lu2023pinat}. While topology-aware predictors can improve prediction quality, they introduce additional overhead from graph construction, message passing, and structured inference. Under extremely limited labels, learning high-dimensional architecture embeddings from scratch may also suffer from cold-start difficulty and overfitting.

Another related direction studies the fusion of multiple ZCPs, for example through adaptive weighting~\cite{lee2024az}. However, purely zero-cost fusion often relies on manually designed aggregation rules or search heuristics, and thus cannot directly learn task-dependent relationships between proxy signals and final accuracy. FENAS bridges few-shot prediction and ZCP fusion by representing each architecture with complementary topology-agnostic physical priors and using minimal ground-truth supervision for data-driven integration. With CAST, heterogeneous proxy scores are aligned and transformed into a compact feature space, allowing lightweight meta-learners to adaptively combine expressivity, trainability, connectivity, and capacity cues for robust architecture ranking.

\section{Methodology}

We now detail the main components of FENAS. The overall idea is to convert each architecture into a compact feature manifold using multi-dimensional zero-cost proxy signals, and then train a lightweight meta-predictor for few-shot performance estimation.

\subsection{Overview}
To address the ranking instability of existing Zero-Cost Proxies (ZCPs) across search spaces~\cite{ning2021evaluating,white2021powerful} and the high evaluation cost of traditional NAS, we propose \textbf{FENAS (Few-shot meta-Ensembling Neural Architecture Search)}. As illustrated in Figure~\ref{fig:framework}, FENAS does not rely on high-dimensional graph topology embeddings~\cite{wen2020neural}. Instead, it extracts low-dimensional physical dynamics features from each architecture and learns a model-agnostic predictor with only a small number of labeled architectures.

The pipeline is decoupled into two stages. First, for an architecture $\alpha$, FENAS extracts raw proxy scores from four complementary physical dimensions and reshapes them through CAST:

\begin{equation}
\mathbf{\Phi}(\alpha) = \text{CAST}(\mathbf{z}_{E} \oplus \mathbf{z}_{T} \oplus \mathbf{z}_{C} \oplus \mathbf{z}_{P})
\label{eq:overview}
\end{equation}

where $\alpha$ denotes a candidate architecture; $\mathbf{z}_{E}$, $\mathbf{z}_{T}$, $\mathbf{z}_{C}$, and $\mathbf{z}_{P}$ denote the proxy scores for expressivity, trainability, connectivity, and parameter capacity, respectively; $\oplus$ denotes feature concatenation; $\text{CAST}(\cdot)$ denotes the proposed Copula Alignment and Synergy Transformation operator; and $\mathbf{\Phi}(\alpha)$ is the resulting feature manifold used for prediction.

Second, the constructed feature manifold is fed into a meta-predictor:

\begin{equation}
\hat{\mathbf{y}} = f_{\theta}(\Phi(\alpha))
\label{eq:pred}
\end{equation}

where $f_{\theta}$ denotes a parameterized meta-predictor with learnable parameters $\theta$, and $\hat{\mathbf{y}}$ denotes the predicted performance of architecture $\alpha$. Since CAST concentrates heterogeneous proxy information into a compact feature space, $f_{\theta}$ can be instantiated as either a nonlinear tree ensemble~\cite{breiman2001random,chen2016xgboost} or a simple linear regression model.

\subsection{Pre-Training Feature Engineering}
\label{sec:feature}

The goal of pre-training feature engineering is to map a discrete architecture $\alpha$ into a continuous, low-dimensional, and informative vector $\Phi(\alpha)$. This process contains two steps: extracting raw ZCP features from four complementary physical dimensions and applying CAST for distribution alignment and high-order feature construction.

\noindent\textbf{\textit{1) ZCPs' Features Extraction}}
\label{sec:proxy}
\newline
For each architecture $\alpha$, we extract scalar proxy scores $z_i \in \mathbb{R}$ from four dimensions: expressivity, trainability, connectivity, and capacity.

\textbf{Expressivity} ($\mathbf{z}_{E}$). We adopt NASWOT~\cite{mellor2021neural} to measure the expressive ability of an untrained architecture. Given an input batch $X$, let $\mathbf{K}$ be the binarized kernel matrix derived from activation states. The expressivity score is computed as~\cite{raghu2017expressive}:

\begin{equation}
\mathbf{z}_{E} = \log \det(\mathbf{K} + \epsilon \mathbf{I})
\label{eq:ze}
\end{equation}

where $\mathbf{K}$ denotes the activation-based kernel matrix, $\epsilon$ is a small positive perturbation for numerical stability, and $\mathbf{I}$ is the identity matrix. A larger $\mathbf{z}_{E}$ indicates stronger activation diversity and lower risk of dimensional collapse.

\textbf{Trainability} ($\mathbf{z}_{T}$). We use the trace of the empirical Fisher information matrix~\cite{martens2020new} to describe the smoothness of the loss landscape and convergence behavior~\cite{karakida2019universal}. Let $g_w = \nabla_w \mathcal{L}_{train}$ be the gradient of the training loss with respect to a trainable weight $w$. The trainability score is:

\begin{equation}
\mathbf{z}_{T} = \sum_{w \in \alpha} \|g_{w}\|^{2}
\label{eq:zt}
\end{equation}

where $w$ denotes a trainable weight in architecture $\alpha$, $\mathcal{L}_{train}$ is the training loss, $g_w$ is the corresponding gradient, and $\|\cdot\|$ denotes the Euclidean norm.

\textbf{Connectivity} ($\mathbf{z}_{C}$). We adopt SynFlow~\cite{tanaka2020pruning} to evaluate global connectivity using a data-free path integral. Given a structure-preserving metric $\mathcal{R}$, the connectivity score is:

\begin{equation}
\mathbf{z}_{C} = \sum_{w \in \alpha} |w| \odot \left|\frac{\partial \mathcal{R}}{\partial w}\right|
\label{eq:zc}
\end{equation}

where $w$ is a trainable weight, $|w|$ is its absolute magnitude, $\mathcal{R}$ denotes the structure-preserving SynFlow objective, $\frac{\partial \mathcal{R}}{\partial w}$ is the derivative of $\mathcal{R}$ with respect to $w$, and $\odot$ denotes the Hadamard product. This score reflects whether information can flow through the architecture without structural bottlenecks.

\textbf{Capacity} ($\mathbf{z}_{P}$). Parameter count provides a basic physical constraint on model capacity. We define it as:

\begin{equation}
\mathbf{z}_{P} = \sum_{l=1}^{L} |w_{l}|
\label{eq:zp}
\end{equation}

where $L$ is the total number of layers, $w_l$ denotes the set of trainable parameters in the $l$-th layer, and $|w_l|$ denotes the number of trainable parameters in that layer. The four scores form the raw feature vector $v(\alpha) = [z_E,z_T,z_C,z_P]^T \in \mathbb{R}^{4}$.

\noindent\textbf{\textit{2) Copula Alignment and Synergy Transformation (CAST)}}
\label{sec:cast}
\newline
The raw proxy vector $v(\alpha)$ contains heterogeneous features with different scales and distributions. To make them comparable and learnable, CAST applies \textit{Copula Alignment}~\cite{nelsen2006introduction} followed by \textit{Synergy Transformation}.

We first use the Empirical Cumulative Distribution Function (ECDF)~\cite{van2000asymptotic} to map each raw feature into a percentile rank. For the $k$-th feature, the aligned value is:

\begin{equation}
u_{k} = \hat{F}_{k}(v_{k}) = \frac{1}{|\mathcal{A}_{ref}|} \sum_{\alpha' \in \mathcal{A}_{ref}} \mathbb{I}(v_{k}(\alpha') \leq v_{k}(\alpha))
\label{eq:ecdf}
\end{equation}

where $v_k(\alpha)$ is the raw value of the $k$-th proxy feature for architecture $\alpha$, $\hat{F}_k(\cdot)$ is the empirical cumulative distribution function of the $k$-th feature, $\mathcal{A}_{ref}$ is the reference architecture set, $|\mathcal{A}_{ref}|$ is its size, $\alpha'$ denotes an architecture sampled from $\mathcal{A}_{ref}$, and $\mathbb{I}(\cdot)$ is an indicator function that equals 1 if the condition is true and 0 otherwise. The resulting $u_k \in [0,1]$ represents the percentile rank of $v_k(\alpha)$ among the reference architectures. Applying this to all four dimensions yields the normalized vector $u=[u_E,u_T,u_C,u_P]^T$.

Next, we construct two high-order synergy features. The first feature measures expressivity per unit capacity:

\begin{equation}
p_{E} = \frac{u_{E}}{\log(1 + u_{P})}
\label{eq:pe}
\end{equation}

where $u_E$ is the aligned expressivity score, $u_P$ is the aligned capacity score, and $p_E$ denotes the efficiency feature.

The second feature models the joint effect of trainability and connectivity:

\begin{equation}
p_{S} = u_{T} \times u_{C}
\label{eq:ps}
\end{equation}

where $u_T$ is the aligned trainability score, $u_C$ is the aligned connectivity score, and $p_S$ denotes the synergy feature. Finally, we concatenate the aligned base features and the two synergy features:

\begin{equation}
\Phi(\alpha) = \text{CAST}(v(\alpha)) = u \oplus [p_{E}, p_{S}]^{T} \in \mathbb{R}^{6}
\label{eq:manifold}
\end{equation}

where $v(\alpha)$ is the raw 4-dimensional proxy vector, $u$ is the aligned 4-dimensional vector, $[p_E,p_S]^T$ is the 2-dimensional synergy vector, $\oplus$ denotes concatenation, and $\Phi(\alpha) \in \mathbb{R}^{6}$ is the final characteristic manifold.

\subsection{Meta-Predictor Instantiation}
\label{sec:predictor}

After CAST, FENAS becomes a model-agnostic prediction framework. The meta-predictor $f_\theta$ can be instantiated by either a simple linear model or a nonlinear tree ensemble.

\noindent\textbf{\textit{1) Linear Instantiation.}} The linear version of FENAS is defined as:

\begin{equation}
\hat{y} = f_{\theta}(\Phi(\alpha)) = w^{T} \Phi(\alpha) + b
\label{eq:linear}
\end{equation}

where $\hat{y}$ is the predicted performance, $w$ is the linear weight vector, $b$ is the bias term, and $\Phi(\alpha)$ is the CAST feature vector of architecture $\alpha$.

\noindent\textbf{\textit{2) Non-linear Tree Ensemble (Default Configuration).}} To capture nonlinear interactions among physical dimensions, we use a tree ensemble such as XGBoost~\cite{chen2016xgboost} or random forests~\cite{breiman2001random} as the default predictor:

\begin{equation}
\hat{y} = f_{\theta}(\Phi(\alpha)) = \sum_{m=1}^{M} \gamma_{m} h_{m}(\Phi(\alpha))
\label{eq:ensemble}
\end{equation}

where $M$ is the number of decision trees, $h_m(\cdot)$ is the $m$-th regression tree, $\gamma_m$ is the weight or learning coefficient of the $m$-th tree, and $\Phi(\alpha)$ is the input CAST feature vector.

\noindent\textbf{\textit{3) Point-wise Regression and Loss Function.}} Under the few-shot setting, we use point-wise regression with Mean Squared Error (MSE). Given a few-shot training set $\mathcal{D}_{train}$, the learning objective is:

\begin{equation}
\mathcal{L}(\theta) = \sum_{i \in \mathcal{D}_{train}} (y_{i} - \hat{y}_{i})^{2} + \Omega(\theta)
\label{eq:loss}
\end{equation}

where $\mathcal{L}(\theta)$ is the training loss, $\theta$ denotes all learnable parameters of the predictor, $\mathcal{D}_{train}$ is the few-shot training set, $i$ indexes an architecture in this set, $y_i$ is its ground-truth validation accuracy, $\hat{y}_i$ is the predicted accuracy, and $\Omega(\theta)$ is a regularization term.

The regularization term depends on the predictor instantiation:

\begin{equation}
\Omega(\theta) =
\begin{cases}
\lambda \|w\|^{2} & \text{for Linear Model} \\
\sum_{m=1}^{M} \left(\gamma T_{m} + \frac{1}{2} \lambda \|w_{m}\|^{2}\right) & \text{for Tree Ensemble}
\end{cases}
\label{eq:reg}
\end{equation}

where $\lambda$ is the L2 regularization coefficient, $w$ is the linear model weight vector, $M$ is the number of trees, $T_m$ is the number of leaf nodes in the $m$-th tree, $w_m$ is the vector of leaf weights in the $m$-th tree, and $\gamma$ is the tree-complexity regularization coefficient. Here $\gamma$ in Eq.~\eqref{eq:reg} is a regularization coefficient and is different from $\gamma_m$ in Eq.~\eqref{eq:ensemble}, which denotes the coefficient of the $m$-th tree.

\section{Experiments}

We evaluate FENAS from three perspectives: few-shot ranking quality, end-to-end architecture search, and the contribution of individual design components. The experiments cover NAS-Bench-101, NAS-Bench-201, and the DARTS search space, and are designed to verify whether the proposed CAST-based proxy fusion is more robust than individual ZCPs and more efficient than topology-heavy predictors.

\subsection{Experimental Settings}

\paragraph{Benchmarks and protocols.}
NAS-Bench-101~\cite{ying2019bench} contains 423,624 CIFAR-10~\cite{krizhevsky2009learning} architectures with diverse graph topologies, while NAS-Bench-201~\cite{Dong2020NAS-Bench-201:} contains 15,625 operation-on-edge architectures with complete records of CIFAR-10, CIFAR-100~\cite{krizhevsky2009learning}, and ImageNet16-120~\cite{chrabaszcz2017downsampled}. We also evaluate practical search in the large-scale DARTS space~\cite{liu2018darts}, an open-domain search space that contains approximately $10^{18}$ candidate architectures. FENAS uses a lightweight Random Forest meta-learner over four CAST-processed ZCPs: Fisher, NASWOT, SynFlow, and \#Params. We compare it with SOTA few-shot predictors such as CAP~\cite{ji2024cap} and PINAT~\cite{lu2023pinat}, as well as pure ZCP baselines. Ranking is measured by Kendall's $\tau$ and Spearman's $\rho$; search quality is measured by top-1 test accuracy and search cost. To ensure statistical robustness, all results are averaged over 10 independent runs with fixed random seeds, and we report both mean and standard deviation.

\begin{table}[t]
\centering
\caption{Ranking results (Kendall's $\tau$) of few-shot predictors.}
\label{tab:nb101_nb201_fewshot}
\vspace{-0.4em}
\begingroup
\setlength{\tabcolsep}{1.8pt}
\renewcommand{\arraystretch}{0.92}
\small
\resizebox{\textwidth}{!}{%
\begin{tabular}{@{}llccccccc@{}}
\toprule
\multirow{2}{*}{Method} & \multirow{2}{*}{Backbone}
& \multicolumn{3}{c}{NB201-C10 ($\tau$)}
& \multicolumn{4}{c}{NB101 ($\tau$)} \\
\cmidrule(lr){3-5} \cmidrule(lr){6-9}
& & \textbf{0.5\%} & \textbf{1\%} & \textbf{3\%}
& \textbf{0.02\%} & \textbf{0.04\%} & \textbf{0.1\%} & \textbf{1\%} \\
\midrule
NP\cite{dudziak2020brp} & \textbf{GCN} & 0.343 & 0.413 & 0.584 & - & - & - & - \\
NAO\cite{luo2018neural} & \textbf{LSTM} & 0.467 & 0.493 & 0.470 & - & - & - & - \\
Arch2Vec\cite{yan2020does} & \textbf{GIN} & 0.542 & 0.573 & 0.601 & 0.435 & 0.511 & 0.561 & 0.596 \\
NPENAS\cite{wen2020neural} & \textbf{GCN} & 0.565 & 0.594 & 0.642 & - & - & - & - \\
TNASP\cite{lu2021tnasp} & \textbf{Trans.} & - & - & - & 0.600 & 0.669 & 0.752 & 0.820 \\
PINAT\cite{lu2023pinat} & \textbf{Trans.} & - & - & - & \textbf{0.679} & \textbf{0.715} & \textbf{0.801} & - \\
CAP\cite{ji2024cap} & \textbf{GIN} & 0.600 & 0.684 & 0.755 & 0.656 & 0.709 & 0.791 & \textbf{0.831} \\
DCL-ENAS\cite{zhang2025evolutionary} & \textbf{GCN} & 0.610 & 0.691 & 0.757 & - & - & - & - \\
\midrule
FENAS (Ours) & \textbf{ZCP+Meta} & \textbf{0.695} & \textbf{0.731} & \textbf{0.757} & 0.616 & 0.641 & 0.658 & 0.683 \\
\bottomrule
\end{tabular}%
}
\endgroup
\vspace{-0.8em}
\end{table}

\subsection{Ranking Performance}
\noindent\textbf{\textit{1) Comparison with Few-shot Methods on NAS-Bench-201.}}
\newline
NAS-Bench-201~\cite{Dong2020NAS-Bench-201:} provides a controlled search space with fixed macro topology, where performance differences are mainly determined by operator-level behaviors. Following standard protocols, we evaluate all methods with 78 (0.5\%), 156 (1\%), and 469 (3\%) labeled samples. As shown in Table~\ref{tab:nb101_nb201_fewshot}, FENAS achieves the best or highly competitive performance across all regimes, with particularly strong gains under extreme few-shot supervision. With only 78 samples, FENAS reaches a $\tau$ of \textbf{0.695} on CIFAR-10, surpassing DCL-ENAS~\cite{zhang2025evolutionary} (0.610) by \textbf{13.9\%} and CAP~\cite{ji2024cap} (0.600) by \textbf{15.8\%}. This advantage is mainly due to the inductive bias of FENAS: instead of learning topology embeddings from scarce labels, it directly exploits physically grounded initialization priors, including expressivity, trainability, and connectivity, through a lightweight meta-ensembling framework. Such a design is well matched to NAS-Bench-201, where the topology is fixed and discriminative information lies primarily in operator-level behavior. Consequently, FENAS can use limited supervision more efficiently and produce more stable rankings than graph-based few-shot predictors~\cite{yan2020does,wen2020neural}.

\noindent\textbf{\textit{2) Comparison with Few-shot Methods on NAS-Bench-101.}}
\newline
NAS-Bench-101~\cite{ying2019bench} is more challenging because of its diverse graph topologies. As shown in Table~\ref{tab:nb101_nb201_fewshot}, FENAS remains competitive in the extremely low-budget regime. With only 100 (0.02\%) labeled architectures, it achieves a $\tau$ of \textbf{0.616}, outperforming Arch2Vec~\cite{yan2020does} (0.435) and TNASP~\cite{lu2021tnasp} (0.600). With 172 (0.04\%) samples, FENAS further reaches \textbf{0.641}, maintaining clear advantages over lightweight representation-based methods. Although some topology-aware predictors such as PINAT~\cite{lu2023pinat} (0.679) obtain higher correlations, these results can still clarify that FENAS can extract useful ranking signals from limited supervision even when structural diversity increases.

This robustness is enabled by the model-agnostic design of FENAS. For NAS-Bench-101, the CAST module adapts feature transformation to better handle heterogeneous architectures while avoiding explicit graph construction and costly message passing. As a result, FENAS keeps low inference complexity and evaluates architectures at approximately \textbf{166.7 ms/arch}, making it suitable for large-scale ranking in practical NAS scenarios. Overall, FENAS provides an efficient and scalable alternative to topology-heavy few-shot predictors.

\noindent\textbf{\textit{3) Comparison with Pure Zero-Cost Proxies.}}
\newline
Since FENAS is built upon zero-cost proxies (ZCPs), we further compare it with directly using individual proxies. As shown in Table~\ref{tab:nb101_nb201_zcp}, single ZCPs show clear performance ceilings and strong instability across search spaces. On NAS-Bench-101~\cite{ying2019bench}, most gradient-based proxies fail to produce reliable rankings: Fisher~\cite{abdelfattah2021zerocost} even gives a negative correlation ($\tau=-0.190$), and SynFlow~\cite{tanaka2020pruning} achieves only $0.250$. Even the strong ZCP-based method NEAR~\cite{husisteinnear} reaches only $0.520$, while FENAS improves the correlation to \textbf{0.658}. This demonstrates that FENAS does not simply reuse raw proxy scores, but converts weak and noisy proxy signals into more reliable ranking predictions through meta-level integration.

%插入表2
\begin{table}[t]
\centering
\caption{Zero-cost ranking performance. FENAS is trained with 424 labels on NB101 and 156 labels on NB201; other ZCPs are training-free. $^*$ marks FENAS base features.}
\label{tab:nb101_nb201_zcp}
\vspace{-0.5em}
\begingroup
\scriptsize
\setlength{\tabcolsep}{2.6pt}
\renewcommand{\arraystretch}{0.86}
\begin{tabular}{@{}lccccccc@{}}
\toprule
\multirow{2}{*}{Method} & NB101 & \multicolumn{2}{c}{NB201-C10} & \multicolumn{2}{c}{NB201-C100} & \multicolumn{2}{c}{NB201-IMG16} \\
\cmidrule(lr){2-2}\cmidrule(lr){3-4}\cmidrule(lr){5-6}\cmidrule(lr){7-8}
& $\tau$ & $\rho$ & $\tau$ & $\rho$ & $\tau$ & $\rho$ & $\tau$ \\
\midrule
Fisher$^*$~\cite{abdelfattah2021zerocost} & -0.190 & 0.477 & 0.346 & 0.527 & 0.381 & 0.474 & 0.344 \\
SynFlow$^*$~\cite{tanaka2020pruning} & 0.250 & 0.700 & 0.516 & 0.703 & 0.524 & 0.645 & 0.474 \\
\#Params$^*$ & 0.300 & 0.753 & 0.578 & 0.728 & 0.553 & 0.692 & 0.520 \\
NASWOT$^*$~\cite{mellor2021neural} & 0.174 & 0.600 & 0.433 & 0.605 & 0.435 & 0.654 & 0.475 \\
NEAR~\cite{husisteinnear} & 0.520 & 0.880 & 0.700 & 0.870 & 0.690 & 0.840 & 0.660 \\
\midrule
FENAS (Ours) & \textbf{0.658} & \textbf{0.907} & \textbf{0.731} & \textbf{0.906} & \textbf{0.730} & \textbf{0.889} & \textbf{0.716} \\
\bottomrule
\end{tabular}
\endgroup
\vspace{-0.8em}
\end{table}

A similar pattern is observed on NAS-Bench-201~\cite{Dong2020NAS-Bench-201:}. Although some ZCPs perform well on specific datasets, their rankings are not consistently reliable across tasks. For example, SynFlow~\cite{tanaka2020pruning} reaches a relatively high correlation on CIFAR-10~\cite{krizhevsky2009learning} ($\rho=0.77$), but degrades on more complex datasets. In contrast, FENAS achieves consistently strong results, reaching \textbf{$\rho=0.907$} and \textbf{$\tau=0.731$} on CIFAR-10, while maintaining high correlations of \textbf{$\tau=0.730$} and \textbf{$\tau=0.716$} on CIFAR-100~\cite{krizhevsky2009learning} and ImageNet16-120~\cite{chrabaszcz2017downsampled}, respectively. These results show that the improvement of FENAS comes from integrating complementary physical perspectives rather than relying on any single proxy. A more detailed analysis of this collaborative effect is provided in Section~4.4.

\subsection{Architecture Search Performance}
After verifying FENAS's ranking accuracy under extreme few-shot settings, we further examine whether such ranking advantages translate into practical architecture search gains. We evaluate FENAS in both a constrained closed-domain space, NAS-Bench-201~\cite{Dong2020NAS-Bench-201:}, and a large-scale open-domain space, DARTS~\cite{liu2018darts}.

\noindent\textbf{\textit{1) Closed-Domain Search on NAS-Bench-201.}}
\newline
For fair comparison with existing SOTA predictors, we randomly sample 50 architectures, about 0.3\% of the search space, with ground-truth accuracies to train the FENAS meta-predictor. The trained predictor scores all 15,625 architectures, and we report the true best test accuracy among the top-50 candidates over 10 random seeds.

As shown in Table~\ref{tab:nb201_search}, FENAS achieves a favorable efficiency--accuracy trade-off. It completes full-space evaluation in only \textbf{$\sim$1138 seconds}, substantially less than many baselines that require $\sim$3800--12000 seconds (however, CAP~\cite{ji2024cap} achieves lower search latency and slightly higher test accuracy on all three datasets), while still identifying competitive architectures. On CIFAR-10~\cite{krizhevsky2009learning}, FENAS obtains \textbf{94.08\%} test accuracy, close to the best reported result of 94.34\% and the ground-truth optimum of 94.37\%. Similar trends are observed on CIFAR-100~\cite{krizhevsky2009learning} and ImageNet16-120~\cite{chrabaszcz2017downsampled}, demonstrating that FENAS can efficiently discover top-tier architectures without explicit topology modeling.

Since Kendall's $\tau$ measures pairwise ordering quality over the entire search space and considers the best ground-truth architecture contained in the predicted top-50 set. Once a near-optimal architecture is included in this set, further improvements produce only modest changes in final test accuracy.

%插入表3
\begin{table}[t]
\centering
\caption{End-to-end search results on NAS-Bench-201~\cite{Dong2020NAS-Bench-201:}. FENAS uses 50 labeled samples to rank the full space; test accuracy (\%) is reported for the best architecture among the top-50 candidates.}
\label{tab:nb201_search}
\vspace{-0.5em}
\begingroup
\setlength{\tabcolsep}{3.2pt}
\renewcommand{\arraystretch}{0.92}
\small
\begin{tabular}{lcccc}
\toprule
Method & Cost & CF10 Test & CF100 Test & IN16 Test \\
 & (s) & (\%) & (\%) & (\%) \\
\midrule
REA~\cite{real2019regularized}          & $\sim$12000 & 93.92 $\pm$ 0.30 & 71.84 $\pm$ 0.99 & 45.15 $\pm$ 0.89 \\
Arch2Vec-BO~\cite{yan2020does}          & $\sim$12000 & 94.18 $\pm$ 0.24 & 73.37 $\pm$ 0.30 & 46.27 $\pm$ 0.37 \\
TNASP~\cite{lu2021tnasp}                & $\sim$4500  & 94.22 $\pm$ 0.18 & 73.35 $\pm$ 0.21 & 46.31 $\pm$ 0.25 \\
PINAT~\cite{lu2023pinat}                & $\sim$3800  & 94.09 $\pm$ 0.14 & 73.26 $\pm$ 0.19 & \textbf{46.55 $\pm$ 0.72} \\
CAP~\cite{ji2024cap}                    & 15.65       & \textbf{94.34 $\pm$ 0.06} & \textbf{73.41 $\pm$ 0.22} & 46.44 $\pm$ 0.36 \\
\midrule
FENAS (Ours) & \textbf{$\sim$1138} & 94.08 $\pm$ 0.21 & 72.13 $\pm$ 0.34 & 46.24 $\pm$ 0.51 \\
\midrule
\textit{Ground Truth} & \textit{-} & \textit{94.37} & \textit{73.51} & \textit{47.31} \\
\bottomrule
\end{tabular}
\endgroup
\vspace{-0.8em}
\end{table}

\noindent\textbf{\textit{2) Search and Retraining on the DARTS Open Domain.}}
\newline
We further evaluate FENAS in the large-scale DARTS search space~\cite{liu2018darts}, which contains approximately $10^{18}$ candidate architectures. Since unconstrained prior-based search may favor degenerate architectures with excessive ``none'' operators~\cite{zelaunderstanding,chu2020fair}, we apply a lightweight structural validity constraint during evolutionary search~\cite{real2019regularized}. In practice, FENAS is fine-tuned using 100 randomly sampled architectures trained for 50 epochs, then used to rank 100,000 valid candidates. The top-3 predicted architectures are retrained from scratch for 600 epochs.

As shown in Table~\ref{tab:darts}, FENAS remains highly efficient, requiring only \textbf{0.3} GPU days while achieving \textbf{97.25\% $\pm$ 0.09\%} test accuracy after retraining. Although FENAS does not outperform PINAT comparing the accuracies, its topology-agnostic design avoids costly graph processing and additional pretraining, enabling direct ranking through lightweight meta-ensembling of initialization-based priors. These results show that FENAS is a practical open-domain NAS method under strict computational budgets.

%插入表4
\begin{table}[t]
\centering
\caption{Search-and-retrain results in the DARTS~\cite{liu2018darts} open domain on CIFAR-10.}
\label{tab:darts}
\vspace{-0.5em}
\begingroup
\setlength{\tabcolsep}{4.0pt}
\renewcommand{\arraystretch}{0.92}
\small
\begin{tabular}{lcccc}
\toprule
\multirow{2}{*}{Method} & Params & Best Acc & Average Acc & Search Cost \\
 & (M) & (\%) & (\%) & (GPU Days) \\
\midrule
DARTS~\cite{liu2018darts}              & 3.3 & -     & 97.24 $\pm$ 0.09 & 4.0 \\
Arch2Vec-BO~\cite{yan2020does}         & 3.6 & 97.52 & 97.44 $\pm$ 0.05 & 100 \\
BANANAS~\cite{white2021bananas}        & 3.6 & -     & 97.33 $\pm$ 0.07 & 11.8 \\
CTNAS~\cite{chen2021contrastive}        & 3.6 & -     & 97.41 $\pm$ 0.04 & \textbf{0.3} \\
PINAT~\cite{lu2023pinat}               & 3.6 & 97.58 & 97.46 $\pm$ 0.08 & \textbf{0.3} \\
CAP~\cite{ji2024cap}                   & 3.3 & \textbf{97.58} & \textbf{97.46 $\pm$ 0.09} & 3.3 \\
\midrule
FENAS (Ours) & 3.4 & 97.34 & 97.25 $\pm$ 0.09 & \textbf{0.3} \\
\bottomrule
\end{tabular}
\endgroup
\vspace{-0.8em}
\end{table}

\subsection{Discussion and Ablation}

\noindent\textbf{\textit{1) Synergy among ZCPs.}}
\newline
To examine how different physical dimensions contribute to ranking performance, we conduct a cumulative feature study on NAS-Bench-201~\cite{Dong2020NAS-Bench-201:} using \textbf{156 labeled architectures} for training and the \textbf{full search space} for evaluation. Following Section~\ref{sec:feature}, the four dimensions --expressivity ($z_E$), trainability ($z_T$), connectivity ($z_C$), and capacity ($z_P$) --are instantiated by NASWOT~\cite{mellor2021neural}, Fisher~\cite{abdelfattah2021zerocost}, SynFlow~\cite{tanaka2020pruning}, and \#Params, respectively.

\begin{figure}[t]
  \centering
  \begin{minipage}[t]{0.36\linewidth}
    \centering
    \includegraphics[width=\linewidth]{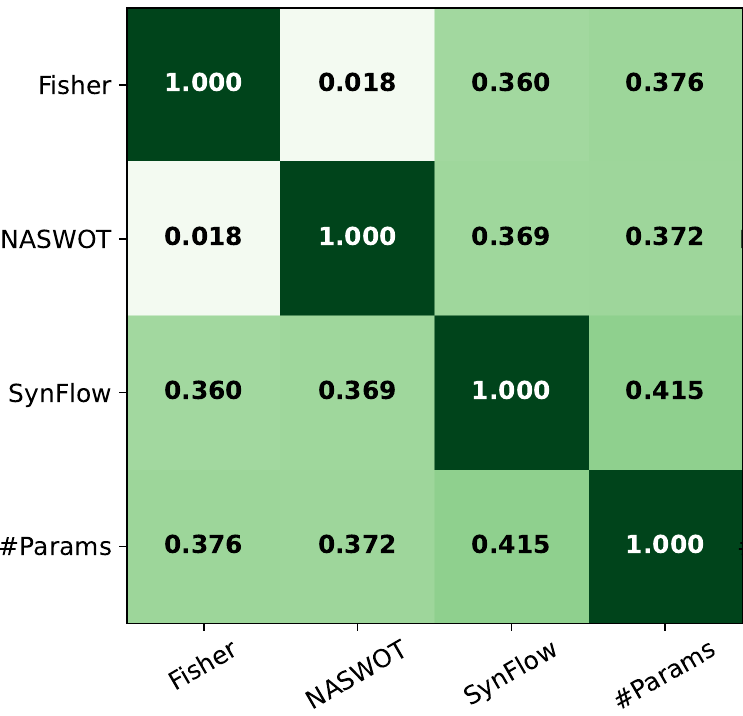}
    \vspace{-0.5em}
    \centerline{\small (a) Proxy correlation}
  \end{minipage}
  \hfill
  \begin{minipage}[t]{0.60\linewidth}
    \centering
    \includegraphics[width=\linewidth]{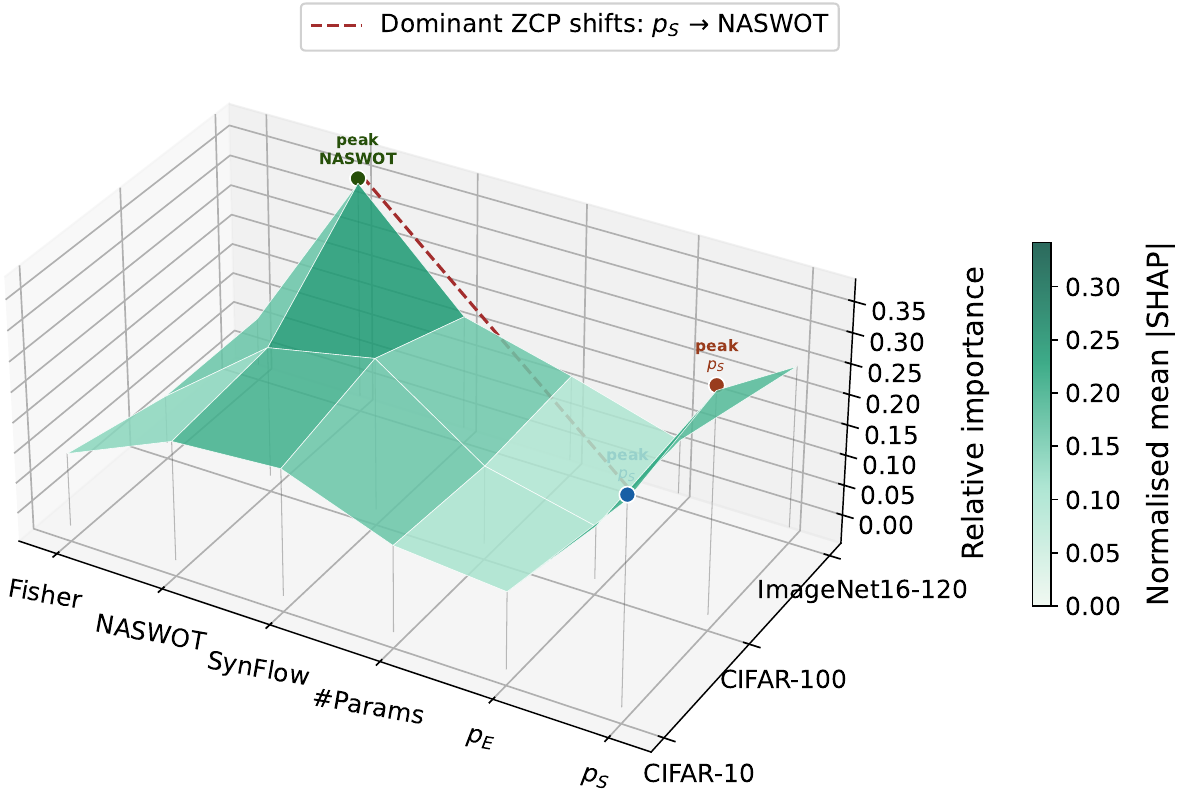}
    \vspace{-0.5em}
    \centerline{\small (b) SHAP importance surface}
  \end{minipage}
  \vspace{-0.4em}
  \caption{Complementarity and adaptive collaboration among ZCPs on NAS-Bench-201.}
  \label{fig:zcp_analysis}
  \label{fig:zcp_corr}
  \label{fig:shap_surface}
  \vspace{-0.8em}
\end{figure}

Before analyzing their cumulative effect, we first verify whether these ZCPs provide complementary signals. As shown in Figure~\ref{fig:zcp_analysis}(a), most off-diagonal $\tau$ correlations are low to moderate, indicating that different proxies capture distinct aspects of architecture quality. This complementarity motivates their integration in FENAS rather than relying on a single dominant proxy.

%插入表5
\begin{table}[!b]
\centering
\caption{Ranking performance ($\tau$) of cumulative combinations of the four physical proxy dimensions on NAS-Bench-201.}
\label{tab:ablation_proxies}
\setlength{\tabcolsep}{3.5pt}
\small
\begin{tabular}{lccc}
\toprule
Cumulative Features & CIFAR-10 ($\tau$) & CIFAR-100 ($\tau$) & ImageNet16-120 ($\tau$) \\
\midrule
$z_T$ & 0.598 & 0.611 & 0.581 \\
$z_T + z_E$ & 0.604 & 0.613 & 0.585 \\
$z_T + z_E + z_C$ & 0.656 & 0.682 & 0.669 \\
$z_T + z_E + z_C + z_P$ & \textbf{0.731} & \textbf{0.730} & \textbf{0.716} \\
\bottomrule
\end{tabular}
\end{table}

As shown in Table~\ref{tab:ablation_proxies}, ranking accuracy improves consistently as more dimensions are introduced. Starting from $z_T$ alone, the predictor achieves $\tau$ values of 0.598/0.611/0.581 on CIFAR-10/100/IN16-120, while using all four dimensions further improves the results to \textbf{0.731}, \textbf{0.730}, and \textbf{0.716}. This steady gain confirms that the selected physical priors are complementary and jointly lead to more stable ranking performance.

We further use SHAP~\cite{lundberg2017unified} to analyze the meta-learner behavior. Figure~\ref{fig:zcp_analysis}(b) shows that FENAS learns adaptive proxy collaboration across datasets instead of relying on a fixed feature. For example, $p_S$ becomes more influential on CIFAR-100, while NASWOT dominates on ImageNet16-120. This task-dependent importance shift explains the cross-dataset robustness of FENAS.

\noindent\textbf{\textit{2) CAST Module.}}
\newline
The CAST module aligns heterogeneous ZCP distributions through ECDF and enhances cross-feature interactions through synergy transformation. We evaluate CAST on NAS-Bench-201/CIFAR-10 using 1\% labeled architectures for meta-training and the full benchmark for testing. Four strategies are compared: raw ZCP features, ECDF-only alignment, Synergy-only, and the complete CAST pipeline.

%插入表6
\begin{table}[!t]
\centering
\caption{Ablation of CAST on NAS-Bench-201/CIFAR-10 with 1\% labels.}
\label{tab:ablation_cast}
\vspace{-0.4em}
\begingroup
\small
\setlength{\tabcolsep}{5.5pt}
\renewcommand{\arraystretch}{0.95}
\begin{tabular*}{0.86\textwidth}{@{\extracolsep{\fill}}lcccc@{}}
\toprule
\multirow{2}{*}{Feature Strategy} & \multicolumn{2}{c}{LR} & \multicolumn{2}{c}{RF} \\
\cmidrule(lr){2-3} \cmidrule(lr){4-5}
& $\tau$ & $\rho$ & $\tau$ & $\rho$ \\
\midrule
$S_1$ (no CAST)   & 0.491 & 0.686 & 0.718 & 0.887 \\
$S_2$ (ECDF only) & 0.545 & 0.735 & 0.729 & 0.895 \\
$S_3$ (Synergy only) & 0.515 & 0.695 & \textbf{0.733} & 0.896 \\
$S_4$ (full CAST) & \textbf{0.560} & \textbf{0.750} & 0.731 & \textbf{0.904} \\
\bottomrule
\end{tabular*}
\endgroup
\vspace{-0.8em}
\end{table}

To verify both effectiveness and model-agnosticity, we test CAST with linear regression (LR) and random forest (RF). As shown in Table~\ref{tab:ablation_cast}, LR benefits substantially from ECDF alignment, improving $\tau$ from 0.491 to 0.545, while the full CAST pipeline further increases it to 0.560. For RF, the gain is more modest. The synergy-only variant obtains a slightly higher $\tau$ of 0.733, whereas full CAST achieves the best $\rho$ of 0.904. These results suggest that CAST is most beneficial for simpler learners and provides a smaller but consistent refinement for the default nonlinear predictor.

\section{Conclusion}

In this paper, we presented FENAS, a topology-agnostic few-shot NAS predictor that integrates multi-dimensional zero-cost proxies with the CAST feature transformation pipeline. By aligning heterogeneous proxy distributions and modeling their complementary interactions, FENAS achieves accurate and stable architecture ranking under extremely limited supervision. Experiments on NAS-Bench-101, NAS-Bench-201, and the DARTS search space show that FENAS provides competitive ranking and search performance with low computational cost. These results suggest that physically grounded proxy fusion is an efficient alternative to topology-heavy predictors, especially when only a small number of evaluated architectures are available. 

\bibliographystyle{splncs04}
\bibliography{custom}

\end{document}